\documentclass[11pt,twocolumn,twoside]{IEEEtran}
\usepackage{amsmath}
\usepackage[margin=0.75in,headheight=0.45in]{geometry}
\usepackage[pdftex]{epsfig}
\usepackage{amsfonts}
\usepackage{amssymb}
\usepackage{fancyhdr}
\include{graphicsx}

\begin{document}
\title{\vspace{0.2in}\sc Deep-dust: Predicting concentrations of fine dust in Seoul using LSTM }
\author{Sookyung Kim$^{1}$\thanks{ $^1$Lawrence Livermore National Laboratory, Livermore, CA $^2$ Korea Institute of Science and Technology, Seoul, Korea}, Jungmin M. Lee$^{1}$, Jiwoo Lee$^{1}$, Jihoon Seo$^{2}$}

\maketitle
\thispagestyle{fancy}
\begin{abstract}
Polluting fine dusts in South Korea which are mainly consisted of biomass burning and fugitive dust blown from dust belt is significant problem these days. Predicting concentrations of fine dust particles in Seoul is challenging because they are product of complicate chemical reactions among gaseous pollutants and also influenced by dynamical interactions between pollutants and multiple climate variables.
Elaborating state-of-art time series analysis techniques using deep learning, non-linear interactions between multiple variables can be captured and used to predict future dust concentration. In this work, we propose the LSTM based model to predict hourly concentration of fine dust at target location in Seoul based on previous concentration of pollutants, dust concentrations and climate variables in surrounding area.
Our results show that proposed model successfully predicts future dust concentrations at 25 target districts(Gu) in Seoul.
\end{abstract}
\begin{figure}
\begin{center}
\epsfxsize=0.97\hsize \epsfbox{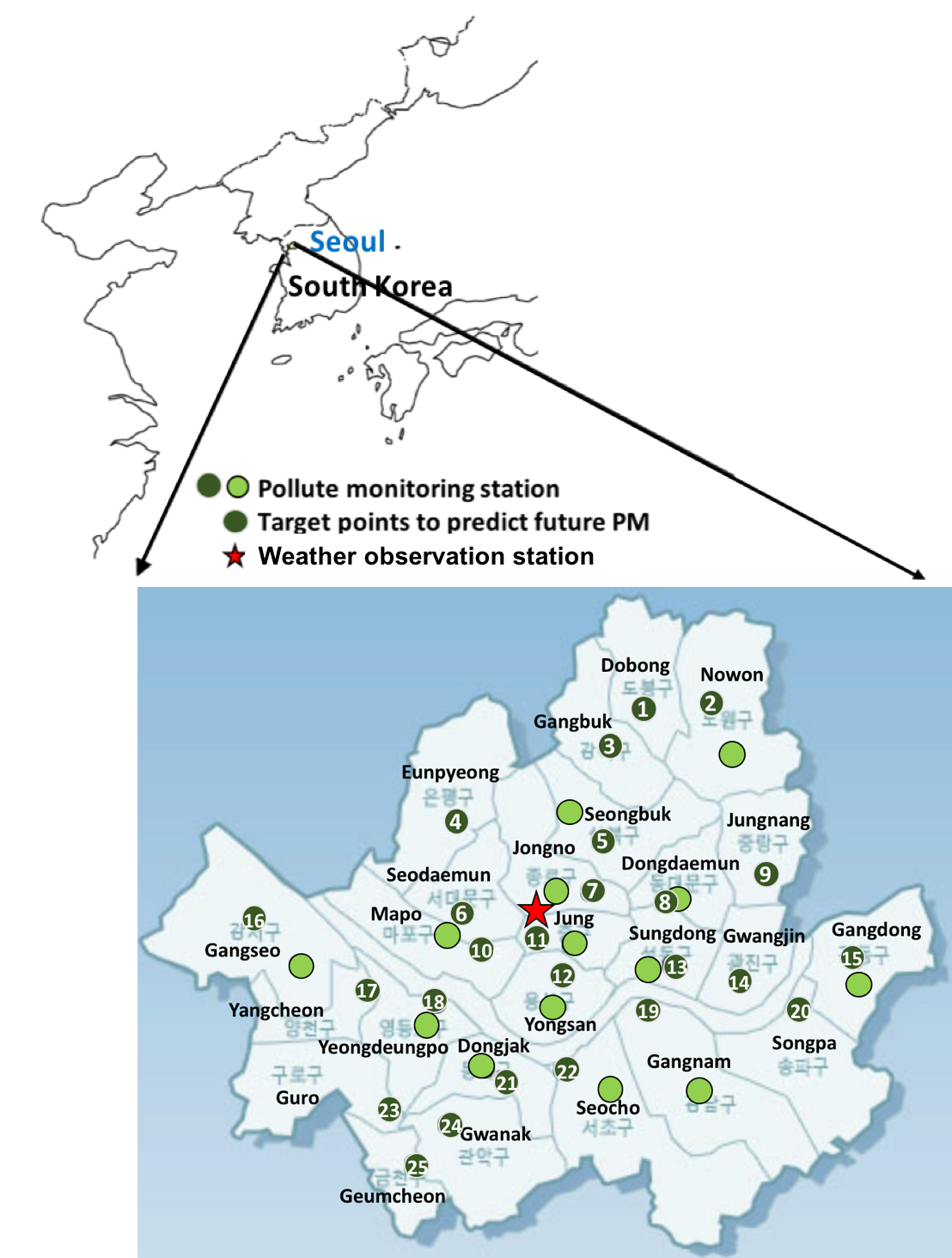}
\end{center}
\caption{Pollute monitoring and climate observation stations in Seoul, Korea.}
%
%
\label{1_station.fig}
\end{figure}
\section{Motivation}
Deteriorating air quality due to polluting fine dusts is significant problem in Korea. 
Located downwind of the prevailing westerlies in the eastern area of the Asian continent, South Korea is significantly influenced by biomass burning and fugitive dust represented from Asian dust.~\cite{duan2004identification,giglio2006global}
Since Korea is a highly populated and industrialized country located in the continental outflow region, its air quality is significantly influenced by both local emission sources and regional transport from remote sources.
The common entity to measure the concentration of fine dust is particulate matter($PM_{10:2.5}$)~\cite{putaud2004european}, and it is reported that the annual average of $PM_{10}$ in Korean peninsula is rapidly and continuously increasing since 1995. According to the report in 2009, the level of $PM_{10}$ in Seoul is much higher than those in New York and Paris.~\cite{park2010estimates,kim2011long} 
As large portion of fine dusts in Korea are consisted of harmful fugitive dust emissions and chemical pollutants, there are large on-going concerns and efforts to monitor and predict air quality related to fine dust concentration.
However, there exists many variables to understand $PM$ concentration including chemical reactions among gaseous pollutants and interactions between pollutants and meteorological parameter. 
Recent advances in deep learning have led to groundbreaking results with complex, nonlinear prediction functions due to its ability to capture the latent abstraction of massive scale complex data~\cite{chin2000learning,goodfellow2016deep}. Specifically, with long short-term memory networks (LSTMs), It is possible to study multiplicative interactions in data to extract relationships between variables across a time sequence.~\cite{bahdanau2014neural,gers1999learning} In this paper, we propose the model using LSTM to predict fine dusts by capturing nonlinear interactions between multiple pollutants and meteorological variables. We predicted hourly concentration of fine dust at target location in Seoul based on previous concentration of pollutants, dust concentrations and climate variables in surrounding area. With our best knowledge, this is the first work to successfully predict concentration of fine dust using deep learning method.

\section{Dataset}
The goal of our model is predicting future concentration of coarse($PM_{10}$) and fine($PM_{2.5}$) dust particles, whose aerodynamic diameters are in range between 2.5 to 10$\mu m$ and below 2.5$\mu m$. Generation of $PM$ dusts 
Fine dust mainly consisted of gaseous pollutants, such as $SO_2$,$CO$,$NO_2$,$O_3$. Gaseous pollutants are divided by two type-(1) primary pollutants ($SO_2$,$CO$,$NO_2$) which are emitted from source, and (2) secondary pollutant ($O_3$) which is generated by photo-chemical reactions interacting with other primary pollutants and climate/meteorological variables. 
Therefore, it is reasonable to assume that there exist nonlinear relation between concentration of future fine dusts and three variables in previous time steps including concentration of 4 gaseous pollutants, concentration of fine dusts and meteorological variables.
In Seoul, there are 39 stations monitoring hourly concentration of 4 gaseous pollutants and fine dusts and 1 climate observation station monitoring hourly meteorological variables.
Figure~\ref{1_station.fig} shows locations of stations.

\section{Model}
\subsection{Featurization}
The model predicts hourly concentration of fine dusts at target location in Seoul based on previous concentration of pollutants, fine dust concentrations and climate variables in surrounding area. There are 39 pollutant stations and 1 climate observation station in Seoul. First, we collected pollutants vector, $P_i$, at $i-th$ pollutant station defined as $P_i$=[$p_{i,j}^t$ for $j$ in $num\_of\_elements$] including concentration of 6 elements at $i-th$ station with following order, $SO_2$, $CO$, $NO_2$, $O_3$, $PM_{10}$, $PM_{2.5}$. Second, we collected climate vector, $C$, at the climate observation station defined as $C$=[$c_{j}^t$ for $j$ in $num\_of\_climate\_variables$] including 9 measured variables with following order,
%
%
%
%
%
%
%
wind speed($m/s$), wind direction(16Directions), humidity($\%$), water vapor pressure($hPa$), dew point temperature($°C$), surface pressure($hPa$), sunlight($hr$), range of vision($m$), surface temperature($°C$). Thirdly, we generated input feature, $\mathbf{x}_{t}$, by concatenating $P_i$ for all 39 pollutant stations with $C$. Formally, $\mathbf{x}_{t}$ can be defined as following,
%
%
$\mathbf{x}_{t} = [concatenate(P_{0:38}, C)]^T =[p_{0,0:5}^t,p_{1,0:5}^t,..,p_{38,0:5}^t,c_{0:9}^t]^T$. 
The output feature, $\mathbf{y}_{t}$ ,includes concentrations of two fine dusts, $PM_{10}$ and $PM_{2.5}$. Formally, $\mathbf{y}_{t}$ can be defined as following, $\mathbf{y}_{t} = [p_{i,10}^t, p_{i,2.5}^t]$.
In this study, we designed model to predict output based on previous $T$ time steps. Therefore, the input of model is time series input feature, $[\mathbf{x}_{0},...,\mathbf{x}_{T-1}]$, and output of model is $\mathbf{y}_{T}$. We discussed $T$ should be multiple of 24 (24 hours, one day) because the chemical reaction to generate secondary pollutants highly affected by sunlight which makes strong correlation between concentration of secondary pollutants with diurnal cycle. In this paper, we took input of previous 48 hours, $[\mathbf{x}_{0},...,\mathbf{x}_{47}]$ , to predict output of next time step, $\mathbf{y}_{48}$ ($i.e: T=48$). All variables in input and output features are normalized within same type of variable to be in a range between 0 and 1.
%
%
\begin{figure}
\begin{center}
\epsfxsize=0.97\hsize \epsfbox{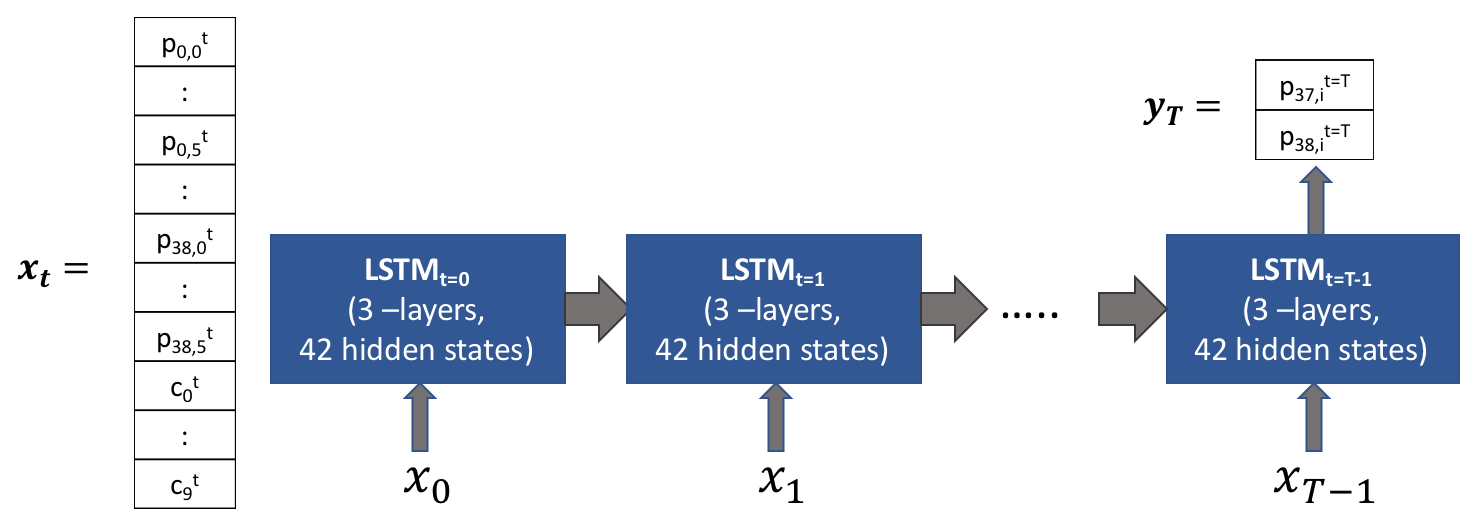}
\end{center}
\caption{Model to predict fine dusts at i-th pollute monitoring station}
\label{3_model}
\end{figure}
\begin{figure*}
\begin{center}
\epsfxsize=0.97\hsize \epsfbox{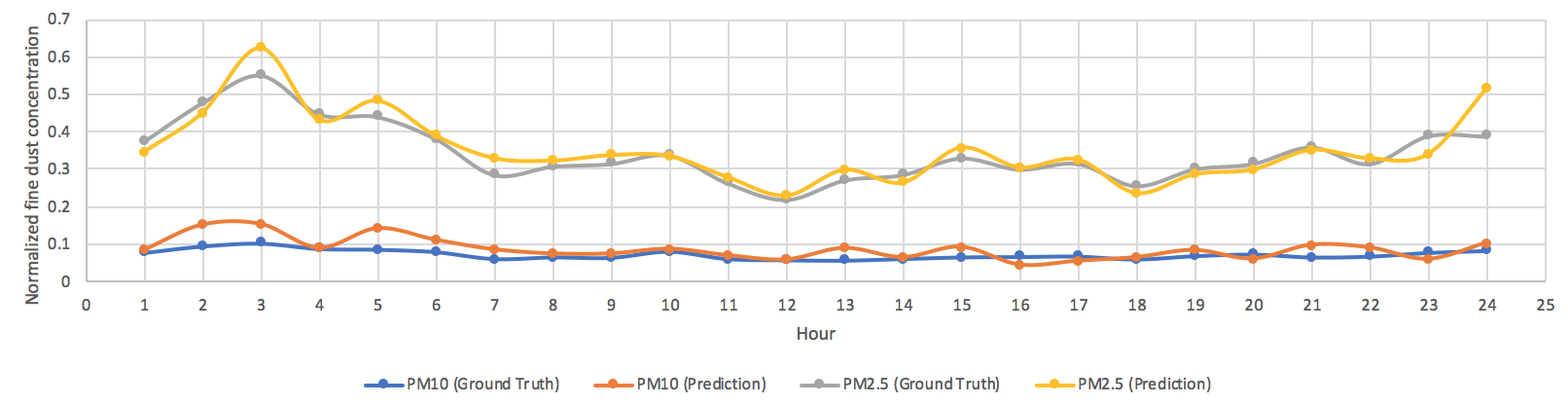}
\end{center}
\caption{Comparison between hourly variation of predicted $PM_{10}$ and $PM_{2.5}$ with ground truth values on sampled day (2017/03/17) at station6 (Seodaemun district).}
%
%
\label{4_result}
\end{figure*}
\begin{table*}
  \centering
  \caption{MSE of models for 25 locations (Unit: 1e-5)}
  \begin{tabular}{c|c|c|c|c|c|c|c|c|c|c|c|c|c} \hline 
    Station &1&2&3&4&5&6&7&8&9&10&11&12&13 \\ \hline
    MSE &$30.40$ & $26.62$ & $29.32$ & $44.09$ & $26.41$& $28.96$& $46.41$& $45.26$& $43.48$& $31.71$ & $44.32$& $27.37$& $28.71$\\ \hline \hline
    Station &14&15&16&17&18&19&20&21&22&23&24&25&  \\ \hline
    MSE &  $46.24$ & $26.27$ & $31.67$ & $31.59$ & $35.51$& $23.24$& $20.35$& $31.71$& $29.29$& $47.86$ & $33.57$& $32.88$& $$ \\ \hline
  \end{tabular}
  \label{table}
\end{table*}
\subsection{Model}
Although input covers variables from all stations, the LSTM weight is designed to be optimized to predict dust concentration at only one target location. 
Therefore, we choose 25 different stations which belong to distinctive districts(Gu) in Seoul (25 stations are numbered and denoted in Figure~\ref{1_station.fig}), and trained 25 distinctive models to predict dust concentration of according locations. 
This is because the size of hidden state to represent information to predict output for every stations should be large enough to potentially demanding large computing cost for training and making difficulty for sub-parameter tunning.
%
%
%
Figure ~\ref{3_model} shows architecture of our fine dust prediction model using LSTM.  Our model $f(\mathbf{x_{0:T-1}};\Theta)$ consists of three-layered LSTM with input-to-state and state-to-state feature with size of $42$. After sequentially feeding in $T$-time series input, $x_{0:T-1}$, output of next time step, $f(\mathbf{x_{0:T-1}};\Theta)$, is obtained from fully connected layer with size of $42 \times 2$  which accepting hidden state at time step $T-1$. $\Theta$ is a set of parameters to optimize. We minimize the element-wise mean squared loss between the output  $f(\mathbf{x_{0:T-1}};\Theta)$, and the ground-truth, $\mathbf{y}_{T}$:
%
%
%
%
%
%
\begin{equation}
  L(\Theta) = \|f(\mathbf{x}_{0:T-1}; \Theta) - \mathbf{y}_T\|_2^2,
\end{equation}

\section{Results}

As the evaluation metric, we report \textbf{mean squared error} between the predicted concentration of $PM_{10}$, $PM_{2.5}$ and ground truth. As shown in TABLE~\ref{table}, mean squared errors for all models for 25 locations are below 45$\times 10^{-5}$ which means less than 10.7 $\%$ error for prediction of each $PM$ value. We also plot the hourly variation of predicted $PM_{10}$ and $PM_{2.5}$ values during the sampled day (3/17/2017) at station6 (Seodaemun district) in test set and compared them with ground truth. As shown in Figure~\ref{4_result}, the predicted normalized $PM_{10}$ and $PM_{2.5}$ values successfully follow the global and local trend of ground truth. From quantitative analysis with mean square error for every 25 locations, and qualitative analysis to compare time series plot of prediction with time series plot of ground truth, we conclude that our model successfully predict future dust concentration in Seoul, Korea. We discussed that future directions of work to improve accuracy of the proposed dust prediction LSTM model would be dividing grid across the predicting area (region in Seoul) and develop spatio-temporal feature to consider spatial correlation between each observation stations by applying ConvLSTM or Graphical Convolutional Neural Network.

\bibliographystyle{ieeetr}
\bibliography{ci_references}

\end{document}